\documentclass[conference]{IEEEtran}
\IEEEoverridecommandlockouts
\usepackage{cite}
\usepackage{amsmath,amssymb,amsfonts}
\usepackage{algorithmic}
\usepackage{subcaption}
\usepackage{graphicx}
\usepackage{textcomp}
\usepackage{pgfplots}
\usepgfplotslibrary{fillbetween}
\pgfplotsset{compat=1.18}
\usepackage{xcolor}
\def\BibTeX{{\rm B\kern-.05em{\sc i\kern-.025em b}\kern-.08em
    T\kern-.1667em\lower.7ex\hbox{E}\kern-.125emX}}
\begin{document}

\title{Gaussian process surrogate model to approximate power grid simulators - An application to the certification of a congestion management controller}

\author{\IEEEauthorblockN{1\textsuperscript{st} Pierre HOUDOUIN}
\IEEEauthorblockA{\textit{RTE} \\
PARIS, FRANCE \\
pierre.houdouin@centralesupelec.fr}
\and
\IEEEauthorblockN{3\textsuperscript{rd} Lucas SALUDJIAN}
\IEEEauthorblockA{\textit{RTE} \\
PARIS, FRANCE \\
lucas.saludjian@rte-france.com}
}

\maketitle

\begin{abstract}
With the digitalization of power grids, physical equations become insufficient to describe the network's behavior, and realistic but time-consuming simulators must be used. Numerical experiments, such as safety validation, that involve simulating a large number of scenarios become computationally intractable. A popular solution to reduce the computational burden is to learn a surrogate model of the simulator with Machine Learning (ML) and then conduct the experiment directly on the fast-to-evaluate surrogate model. Among the various ML possibilities for building surrogate models, Gaussian Processes (GPs) emerged as a popular solution due to their flexibility, data efficiency, and interpretability. Their probabilistic nature enables them to provide both predictions and Uncertainty Quantification (UQ). This paper starts with a discussion on the interest of using GPs to approximate power grid simulators and speed up numerical experiments. Such simulators, however, often violate the GP's underlying Gaussian assumption, leading to poor approximations. To address this limitation, an approach that consists in adding an adaptive residual uncertainty term to the UQ is proposed. It enables the GP to remain accurate and reliable despite the simulator's non-Gaussian behaviors. This approach is successfully applied to the certification of the proper functioning of a congestion management controller in a static use case.
\end{abstract}
 

\section{Introduction}

The energy transition complicates congestion management. With the large-scale integration of Renewable Energy Sources (RES), production patterns evolve, becoming more variable and unpredictable. This poses operational challenges \cite{Meyer2020Power} for Transmission System Operators (TSOs) as the lines' power flows can no longer be anticipated as before. To avoid overly conservative RES production limitations, congestion management must now be performed as close to real-time as possible, driving the need for automation.

To overcome this challenge and automate congestion management, the French TSO called RTE (Réseau de Transport d'Electricité) developed the NAZA (New Adaptive Zonal Automaton) \cite{Hoang2021Power}, \cite{Hoang2021Predictive}, \cite{Straub2018Zonal} controllers. NAZA are zonal controllers that automatically take close-to-real-time actions to prevent congestion. A Model Predictive Control (MPC) algorithm integrated into each controller runs every 15 seconds to determine the actions to perform. It includes limiting RES production, managing batteries, and performing zonal topological modification. As the controllers are gradually deployed in the French transmission network, with 10 new controllers every year and around 200 expected by 2035, RTE aims to ensure that the MPC does not threaten the grid's safety and effectively avoids congestion.

A way to certify the MPC's proper functioning consists in computing its failure probability \cite{Moss2024Bayesian}. In this case, it comes down to estimating the proportion of scenarios where congestion occurs despite the scenario being considered manageable for the MPC. To determine whether a scenario induced congestion or not, power grid simulators that include the MPC's actions can be used. A naive solution to achieve the certification is then to randomly draw manageable scenarios, determine their outcome with the simulator, and estimate the failure probability when enough simulations have been performed. The problem, however, is that a simulation can last minutes, and there are potentially millions of scenarios to test. Such a brute-force approach is thus computationally intractable in practice.

\vspace{0.2cm}

A popular method in the literature to reduce the computational cost of experiments involving time-consuming simulators is to learn a surrogate model, an approximation, of this simulator with Machine Learning (ML) \cite{Ioss2018Advanced}, \cite{Marrel2008Efficient}, \cite{Marrel2024Probabilistic}, \cite{Bastos2009Diagnostics}, \cite{Carvalho2022Choosing}. The experiment can then be conducted directly on the fast-to-evaluate surrogate model, leading to a huge time gain. However, if the model is inaccurate, trusting its prediction could considerably hinder the experiment's results. To avoid this, an uncertainty-aware surrogate model that is able to assess its confidence is trained. The surrogate model replaces the simulator only when it exhibits enough confidence about its prediction, reducing the risk of trusting a flawed model.

Many approaches exist to build uncertainty-aware machine learning models. Standard ML models can be combined with model-free Uncertainty Quantification (UQ) techniques such as conformal predictions \cite{Shafer2008Tutorial}. Another possibility is to adapt existing prediction models to enable them to quantify uncertainty, such as quantile regression \cite{Koenker2017Quantile}. A third increasingly popular option is to use Gaussian Processes \cite{Rasmussen2004Gaussian} (GPs). GPs are based on a statistical framework that provides a full probability distribution over the possible outcomes, enabling both prediction and UQ. GPs are also \textbf{flexible}, \textbf{data-efficient}, and \textbf{interpretable} ML models. They are increasingly used in many industrial fields to approximate computationally intensive simulators.

This work is the continuation of \cite{Houdouin2024Certification}, where GPs were successfully leveraged to build a certification workflow that estimates the MPC's failure probability, tested on a fictitious simulator. In this paper, a more realistic use case is considered. A simple simulator is still considered, but with the real MPC actions emulated. These actions introduce non-linearities that violate the underlying GPs' Gaussian assumption and are not properly captured by a vanilla GP. A new approach leverages the add-on of an adaptive residual uncertainty term to the Gaussian posterior to preserve the GP's reliability despite the non-Gaussian behaviors. The paper is structured as follows:

\begin{itemize}
    \item Section II contains the first contribution: a discussion on why GPs are well-suited to build uncertainty-aware surrogate models and details
regarding their use to speed up the certification.
    \item Section III presents the second contribution: the new approach with the adaptive residual uncertainty to better deal with non-Gaussian behaviors.
    \item Section IV provides experimental results of the certification workflow tested on a real zone of the French power grid.
    \item Conclusions and perspectives are drawn in section V.
\end{itemize}

\section{Gaussian processes, a popular approach for building surrogate models}

\subsection{Machine-learning techniques to build an accuracy-aware surrogate model}

Learning a surrogate model to approximate complex simulators is a popular technique to accelerate numerical experiments. In many applications, however, trusting a flawed model can have severe consequences. To prevent this, a solution is to learn an uncertainty-aware surrogate model that is able to assess its own confidence. The simulator can then be replaced only when the model is sufficiently confident about its prediction, reducing the risk of trusting an inaccurate model. For this reason, uncertainty quantification has become a golden standard in many industrial applications. It can be achieved in several ways.

A first approach consists in combining standard ML algorithms with model-free UQ techniques. Standard ML algorithms already used in the literature to build surrogate models include gradient boosting methods, polynomial chaos expansion, random forest, response surface methods, and polynomial splines. It can be plugged with model-free UQ techniques such as Monte Carlo, conformal predictions, and various ensemble methods (Bootstrap prediction intervals, infinitesimal Jackknife, and Out-Of-Bag errors). This first approach offers great flexibility in the surrogate model selection, and UQ methods rely on mild assumptions that are often respected. The main drawback of these model-free UQ techniques is that they tend to produce overly large confidence intervals with small datasets. They also have a high computational cost, making them poorly suited to building fast-to-evaluate uncertainty-aware surrogate models.

A second possible approach consists in adapting existing prediction models to enable them to quantify uncertainty. For example, quantile regression (QR) methods modify the loss function so that the machine learning model predicts a specific quantile of the expected output. Predicting the median, the 5\%, and the 95\% quantiles allows obtaining a prediction and the associated 90\% confidence interval. Other works also propose to combine the predictive power of artificial neural networks (ANN) with deep-learning-specific UQ techniques, such as the delta method, the mean-variance estimation, the lower upper bound estimation, and quality-driven ensemble methods. There are two drawbacks to this second approach. First, for both QR and ANN, a lot of data is required to train the models effectively. Second, the obtained surrogate model is black-box, which is undesirable for safety-critical applications such as ensuring power grid security.

A third increasingly adopted solution to build accuracy-aware surrogate models is Gaussian processes. Instead of a simple prediction, GPs provide a full posterior Gaussian distribution $\mathcal{N}\left( \mu_*, \sigma_*^2 \right)$ of the outcome, which is much more informative. The posterior mean $\mu_*$ constitutes the prediction, and the posterior standard deviation $\sigma_*$ is used for UQ. Many reasons explain the growing popularity of GPs.

\begin{itemize}
    \item \textbf{GPs are flexible non-parametric models}. Many machine learning models are parametric. They learn an approximation $\Tilde{f}$ that belongs to a parametric set of functions $\{\Tilde{f}_{\boldsymbol{\theta}}, \boldsymbol{\theta} \in \Theta \}$ (polynomial regressors, neural networks...). The hyperparameters $\boldsymbol{\theta}$ are optimized with the train data that no longer contribute to the model afterward. All the exploited information is compressed into $\boldsymbol{\theta}$. Unlike these models, the GP does not assume a fixed number of parameters to approximate the function. Instead, it automatically adapts the model's complexity to the dataset size, which is very flexible.
    \item \textbf{GPs provide both prediction and UQ}. Instead of looking for one prediction that best fits the data, the GP assumes that many approximations could plausibly model the data. The GP then assigns to each approximation a probability of actually representing the true function. The most probable approximation constitutes the GP prediction; the other likely ones contribute to UQ. In practice, when asked to predict the outcome of a new scenario $\boldsymbol{x_{\text{new}}}$, the GP returns a posterior Gaussian distribution $y_{\text{new}} \sim \mathcal{N}\left( \mu_*, \sigma_*^2 \right)$ that depends on $\boldsymbol{x_{\text{new}}}$. A $1-\alpha$ confidence interval is derived using the posterior standard deviation $\sigma_*$ and the $1-\frac{\alpha}{2}$ Gaussian quantile.
    \begin{align*}
        \mathcal{I}_{1-\alpha} = [\mu_* - q_{1-\frac{\alpha}{2}} \sigma_*, \mu_* + q_{1-\frac{\alpha}{2}} \sigma_*]
    \end{align*}
    \item \textbf{GPs are data efficient}. GPs can provide accurate predictions and narrow confidence intervals with very few samples. They rely on the intuition that close inputs have close outputs. If $\boldsymbol{x_{\text{new}}}$ is close enough to previously simulated scenarios, confident predictions can be obtained even though the dataset is small.
    \item \textbf{GPs are not black-box}, predictions and UQ are interpretable. Given $N$ simulated scenarios $(\boldsymbol{x_1}, y_1),...,(\boldsymbol{x_N}, y_N)$ and a new scenario $\boldsymbol{x_{\text{new}}}$:
    \begin{itemize}
        \item The GP prediction $\mu_* = \sum_{n=1}^N w_n y_n$ is obtained with a weighted linear combination of the observed outputs. The factor $w_n$ depends on $\boldsymbol{x_{\text{new}}}$. It reflects how correlated $\boldsymbol{x_n}$ is with $\boldsymbol{x_{\text{new}}}$ and the propensity of $\boldsymbol{x_n}$ to provide unique information regarding the rest of the dataset. 
        \item The uncertainty $\sigma_*^2 = \sigma_{\text{prior}}^2 - \sigma_{\text{reduction}}^2$ is the difference of the prior uncertainty with a variance reduction term. The more $\boldsymbol{x_n}$ highly correlated with $\boldsymbol{x_{\text{new}}}$ there are, the higher the variance reduction is. Intuitively, this reflects that when there is a lot of observed data close to the new input, the function's behavior is well-characterized, and the uncertainty around the prediction is small.
    \end{itemize}
\end{itemize}

\subsection{Gaussian processes: a practical perspective}

Intuitively, many scenarios do not require a simulation. For example, RES will not create congestion if there is no wind and no sun. The goal of the Gaussian process surrogate model is to replace the simulator for scenarios with obvious outcomes to avoid some simulations and save time. Rather than directly predicting the presence or absence of congestion, the GP predicts the maximum relative charge $y$ observed on the lines of the zone. If $y > 1$, a line is overloaded, and congestion has occurred. If $y \leq 1$, all the lines are safe, and no congestion has occurred. Predicting the maximum relative charge enables transforming a classification problem into a smoother regression problem better handled by a GP. At each workflow iteration, the GP leverages the previous simulations to return the posterior distribution $y_{\text{new}} \sim \mathcal{N}(\mu_*, \sigma_*^2)$ of the maximum relative charge induced by the new scenario $\boldsymbol{x_{\text{new}}}$. According to the GP, the probability for the scenario to create a congestion is $p_{\text{congestion}} = \frac{1}{2} \left[ 1 - erf\left( \frac{1 - \mu_*}{\sqrt{2} \sigma_*} \right) \right]$, $erf$ being the error function \cite{Abramowitz1964Handbook}. The prediction is kept if the GP is sufficiently confident about the presence or absence of congestion, which means that $p_{\text{congestion}}$ is close enough to 0 or 1. This is assessed by comparing the congestion probability to a user-defined confidence threshold $\beta$.

\begin{itemize}
    \item If $p_{\text{congestion}} < \beta$, the GP confidently predicts the absence of congestion, the prediction is kept, and a simulation is avoided.
    \item If $p_{\text{congestion}} > 1 - \beta$, the GP confidently predicts the presence of congestion, the prediction is kept, and a simulation is avoided.
    \item If $\beta < p_{\text{congestion}} < 1 - \beta$, the GP is not confident enough about the scenario's outcome. The simulation is performed, and the new sample improves the GP for future iterations.
\end{itemize}

The simulation is avoided for scenarios that are not supposed to provide additional useful information for the failure probability estimation. In practice, the GP replaces the simulator when $\left| \frac{1 - \mu_*}{\sigma_*} \right| > \sqrt{2} erf^{-1}(1-2\beta)$. This happens in two types of situations:

\begin{itemize}
    \item The predicted maximum relative charge $\mu_*$ is far from 1, the congestion threshold. According to the GP, the scenario is either very likely or very unlikely to create congestion. This is the case of a scenario with no wind, for example.
    \item The posterior standard deviation $\sigma_*$ is very small. This means that although $\mu_*$ is close to the congestion threshold, there are enough simulated scenarios nearby for the GP to be sufficiently confident about the outcome.
\end{itemize}

The success of this surrogate-model-based approach hinges on two key elements. First is the GP's ability to accurately approximate the simulator's behavior and provide reliable predictions. Second, the GP must learn fast to avoid enough simulations and ensure an efficient certification.

\vspace{0.2cm}

Realistic power grid simulators, however, do not always respect the underlying GP's Gaussian assumption, which can hinder the GP's accuracy. An adaptation of the standard GP model to preserve a reliable UQ is proposed in the next part.

\section{Adaptive residual uncertainty to deal with non-Gaussian behaviors}

To illustrate the second contribution of this paper and for the experiments, a fictitious power grid simulator using the load flow equations is implemented. In this section, a simple zone with a single line and a single RES production unit is considered.

\begin{itemize}
    \item Given the production $x \in \mathbb{R}$ and the Power Transfer Distribution Factor $m \in \mathbb{R}$, the line's power flow $y \in \mathbb{R}$ is determined by the linear relationship : $y=m \times x$.
    \item A $min$ function is then introduced to model the MPC production limitation order. Production is limited to ensure that the power flow remains below a user-defined threshold $F_{\text{max}}$: $y = m \times min(x, F_{\text{max}} / m)$.
\end{itemize}

 The simulator's function is plotted in Fig. 1. Although simple and univariate, it is the type of non-linear behavior expected with a realistic power grid simulator when the MPC limits RES production. To facilitate intuition about how the dynamic generalizes with multiple RES units, Fig. 2 illustrates the two-dimensional case.

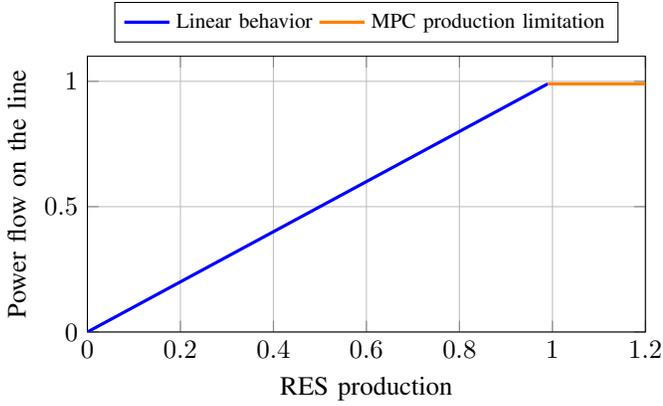
\begin{figure}[htbp]
    \centering
    \begin{tikzpicture}
        \begin{axis}[width=9cm, height=5.25cm, xmin=0, xmax=1.2, ymin=0,   ymax=1.1, xlabel={RES production}, ylabel={Power flow on the line}, grid=major, legend style={at={(0.5,1.05)}, anchor=south, legend columns=2, font=\footnotesize}]
        \addplot [blue, very thick, domain=0:0.99] {x};
        \addlegendentry{Linear behavior}
        \addplot [orange, very thick, domain=0.99:1.2] {0.99};
        \addlegendentry{MPC production limitation}
      \end{axis}
    \end{tikzpicture}
    \caption{Univariate simulator's function}
\end{figure}

\begin{figure}[htbp]
    \centering
        \includegraphics[width=0.45\textwidth]{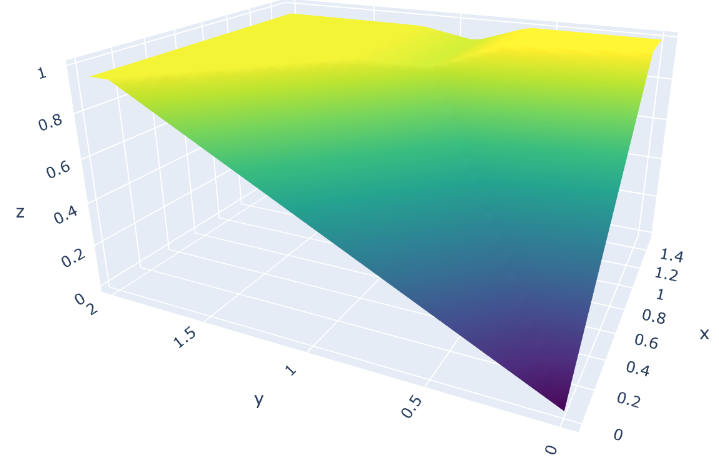}
    \caption{Bivariate simulator's function}
\end{figure}

\subsection{Limitations of the vanilla GP}

Mathematically, a GP is a random function whose realizations are called trajectories. These trajectories have Gaussian behavior: they are smooth and oscillate with similar variation speeds. To build an approximation, the GP interpolates the observed points and then fills the gaps by inferring the most likely Gaussian behavior. The prior Gaussian behavior assumption contributes all the more to the prediction as few observations are available, which grants the GP its data efficiency. 

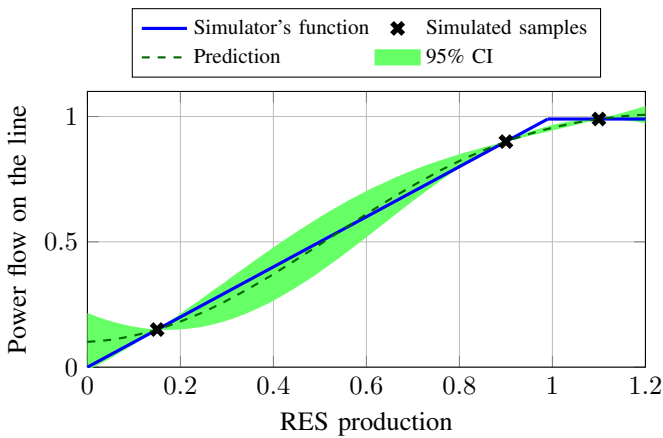
\begin{figure}[htbp]
    \centering
    \begin{tikzpicture}
        \begin{axis}[width=9cm, height=5.25cm, xmin=0, xmax=1.2, ymin=0, ymax=1.1, xlabel={RES production}, ylabel={Power flow on the line}, grid=major, legend style={at={(0.5,1.05)}, anchor=south, legend columns=2, font=\footnotesize}, legend cell align={left}]
        \addplot [blue, very thick, domain=0:0.99] {x};
        \addlegendentry{Simulator's function}
        \addplot [blue, very thick, domain=0.99:1.2, forget plot] {0.99};
        \addplot[only marks, mark=x, line width=2pt, mark size=3pt, black] coordinates {(0.150, 0.150) (0.900, 0.900) (1.100, 0.990)};
        \addlegendentry{Simulated samples}
        \addplot[green!40!black, dashed, thick] coordinates {(0.000, 0.101) (0.024, 0.104) (0.049, 0.109) (0.073, 0.116) (0.098, 0.125) (0.122, 0.136) (0.147, 0.148) (0.171, 0.163) (0.196, 0.179) (0.220, 0.197) (0.245, 0.216) (0.269, 0.237) (0.294, 0.260) (0.318, 0.284) (0.343, 0.309) (0.367, 0.335) (0.392, 0.362) (0.416, 0.390) (0.441, 0.419) (0.465, 0.448) (0.490, 0.477) (0.514, 0.507) (0.539, 0.537) (0.563, 0.566) (0.588, 0.596) (0.612, 0.625) (0.637, 0.654) (0.661, 0.682) (0.686, 0.709) (0.710, 0.735) (0.735, 0.760) (0.759, 0.785) (0.784, 0.808) (0.808, 0.830) (0.833, 0.850) (0.857, 0.869) (0.882, 0.887) (0.906, 0.904) (0.931, 0.919) (0.955, 0.933) (0.980, 0.946) (1.004, 0.957) (1.029, 0.967) (1.053, 0.976) (1.078, 0.984) (1.102, 0.991) (1.127, 0.996) (1.151, 1.001) (1.176, 1.005) (1.200, 1.007)};
        \addlegendentry{Prediction}
        \addplot[green!60, name path=upper, draw=none, forget plot] coordinates {(0.000, 0.216) (0.024, 0.199) (0.049, 0.184) (0.073, 0.172) (0.098, 0.162) (0.122, 0.155) (0.147, 0.150) (0.171, 0.177) (0.196, 0.208) (0.220, 0.240) (0.245, 0.272) (0.269, 0.305) (0.294, 0.337) (0.318, 0.370) (0.343, 0.403) (0.367, 0.435) (0.392, 0.467)(0.416, 0.498) (0.441, 0.529) (0.465, 0.558) (0.490, 0.587) (0.514, 0.615) (0.539, 0.641) (0.563, 0.667) (0.588, 0.691) (0.612, 0.714) (0.637, 0.735) (0.661, 0.756) (0.686, 0.775) (0.710, 0.793) (0.735, 0.810) (0.759, 0.825) (0.784, 0.840) (0.808, 0.854) (0.833, 0.867) (0.857, 0.880) (0.882, 0.891) (0.906, 0.905) (0.931, 0.925) (0.955, 0.942) (0.980, 0.956) (1.004, 0.968) (1.029, 0.978) (1.053, 0.984) (1.078, 0.988) (1.102, 0.991) (1.127, 1.003) (1.151, 1.015) (1.176, 1.028) (1.200, 1.042)};
        \addplot[green!60, name path=lower, draw=none, forget plot] coordinates {(0.000, -0.013) (0.024, 0.010) (0.049, 0.035) (0.073, 0.061) (0.098, 0.088) (0.122, 0.117) (0.147, 0.146) (0.171, 0.149) (0.196, 0.150) (0.220, 0.154) (0.245, 0.160) (0.269, 0.170) (0.294, 0.182) (0.318, 0.197) (0.343, 0.215) (0.367, 0.235) (0.392, 0.257) (0.416, 0.282) (0.441, 0.309) (0.465, 0.337) (0.490, 0.367) (0.514, 0.399) (0.539, 0.432) (0.563, 0.466) (0.588, 0.501) (0.612, 0.536) (0.637, 0.572) (0.661, 0.607) (0.686, 0.643) (0.710, 0.677) (0.735, 0.711) (0.759, 0.744) (0.784, 0.775) (0.808, 0.805) (0.833, 0.833) (0.857, 0.859) (0.882, 0.883) (0.906, 0.903) (0.931, 0.914) (0.955, 0.925) (0.980, 0.935) (1.004, 0.946) (1.029, 0.957) (1.053, 0.968) (1.078, 0.979) (1.102, 0.990) (1.127, 0.989) (1.151, 0.986) (1.176, 0.981) (1.200, 0.973)};
        \addplot[green!60, opacity=1] fill between[of=upper and lower];
        \addlegendimage{area legend, draw=none, fill=green!60, opacity=1}
        \addlegendentry{95\% CI}
        \end{axis}
    \end{tikzpicture}
    \caption{Inaccurate GP approximation near the breakpoint}
\end{figure}

In this use case, the power grid simulator exhibits a non-linearity with a breakpoint at $x = F_{\text{max}} / m$. Such non-linearity is, however, not a Gaussian behavior, which is why the GP produces an inaccurate prediction near the breakpoint, as illustrated in Fig. 3. When more samples are available, the GP gives more importance to the data, and the prior Gaussian assumption becomes less predominant, leading to a better prediction. Fig. 4 highlights the improved GP approximation when more scenarios are simulated near the breakpoint.

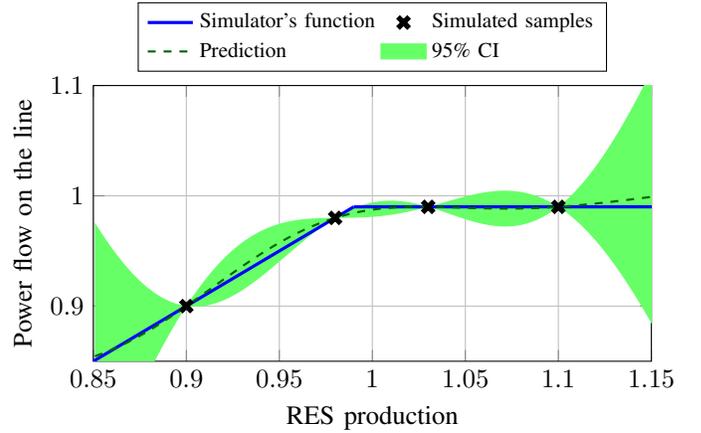
\begin{figure}[htbp]
    \centering
    \begin{tikzpicture}
        \begin{axis}[width=9cm, height=5.25cm, xmin=0.85, xmax=1.15, ymin=0.85, ymax=1.1, xlabel={RES production}, ylabel={Power flow on the line}, grid=major, legend style={at={(0.5,1.05)}, anchor=south, legend columns=2, font=\footnotesize}, legend cell align={left}]
        \addplot [blue, very thick, domain=0:0.99] {x};
        \addlegendentry{Simulator's function}
        \addplot [blue, very thick, domain=0.99:1.2, forget plot] {0.99};
        \addplot[only marks, mark=x, line width=2pt, mark size=3pt, black] coordinates {(0.900, 0.900) (0.980, 0.980) (1.030, 0.99) (1.100, 0.990)};
        \addlegendentry{Simulated samples}
        \addplot[green!40!black, dashed, thick] coordinates {(0.85084,0.85455) (0.85485,0.857) (0.85886,0.85971) (0.86288,0.86268) (0.86689,0.86589) (0.8709,0.86934) (0.87492,0.87301) (0.87893,0.87689) (0.88294,0.88096) (0.88696,0.88521) (0.89097,0.88962) (0.89498,0.89416) (0.899,0.89882) (0.90301,0.90357) (0.90702,0.90839) (0.91104,0.91325) (0.91505,0.91812) (0.91906,0.92299) (0.92308,0.92782) (0.92709,0.93259) (0.9311,0.93728) (0.93512,0.94186) (0.93913,0.9463) (0.94314,0.95059) (0.94716,0.9547) (0.95117,0.95862) (0.95518,0.96233) (0.9592,0.96582) (0.96321,0.96907) (0.96722,0.97209) (0.97124,0.97485) (0.97525,0.97736) (0.97926,0.97961) (0.98328,0.98162) (0.98729,0.98338) (0.9913,0.9849) (0.99532,0.9862) (0.99933,0.98727) (1.00334,0.98815) (1.00736,0.98883) (1.01137,0.98934) (1.01538,0.9897) (1.0194,0.98993) (1.02341,0.99003) (1.02742,0.99004) (1.03144,0.98997) (1.03545,0.98983) (1.03946,0.98965) (1.04348,0.98945) (1.04749,0.98923) (1.05151,0.98902) (1.05552,0.98882) (1.05953,0.98865) (1.06355,0.98852) (1.06756,0.98843) (1.07157,0.9884) (1.07559,0.98842) (1.0796,0.98851) (1.08361,0.98867) (1.08763,0.98889) (1.09164,0.98918) (1.09565,0.98954) (1.09967,0.98996) (1.10368,0.99045) (1.10769,0.99099) (1.11171,0.99159) (1.11572,0.99224) (1.11973,0.99293) (1.12375,0.99366) (1.12776,0.99443) (1.13177,0.99523) (1.13579,0.99605) (1.1398,0.99688) (1.14381,0.99773) (1.14783,0.99858) (1.15184,0.99944)};
        \addlegendentry{Prediction}
        \addplot[green!60, name path=upper, draw=none, forget plot] coordinates {(0.85084,0.97562) (0.85485,0.96506) (0.85886,0.95517) (0.86288,0.94599) (0.86689,0.93757) (0.8709,0.92992) (0.87492,0.92309) (0.87893,0.91711) (0.88294,0.91198) (0.88696,0.90773) (0.89097,0.90436) (0.89498,0.90188) (0.899,0.90028) (0.90301,0.90763) (0.90702,0.91717) (0.91104,0.92599) (0.91505,0.93407) (0.91906,0.94142) (0.92308,0.94802) (0.92709,0.95389) (0.9311,0.95905) (0.93512,0.96351) (0.93913,0.96732) (0.94314,0.9705) (0.94716,0.97311) (0.95117,0.97519) (0.95518,0.9768) (0.9592,0.978) (0.96321,0.97885) (0.96722,0.9794) (0.97124,0.97974) (0.97525,0.97992) (0.97926,0.98004) (0.98328,0.98322) (0.98729,0.98667) (0.9913,0.98959) (0.99532,0.99194) (0.99933,0.99372) (1.00334,0.99491) (1.00736,0.99553) (1.01137,0.9956) (1.01538,0.99516) (1.0194,0.99425) (1.02341,0.99292) (1.02742,0.99125) (1.03144,0.99069) (1.03545,0.99258) (1.03946,0.99452) (1.04348,0.99645) (1.04749,0.99831) (1.05151,1.00004) (1.05552,1.00159) (1.05953,1.00289) (1.06355,1.00387) (1.06756,1.0045) (1.07157,1.00469) (1.07559,1.00441) (1.0796,1.0036) (1.08361,1.00223) (1.08763,1.00024) (1.09164,0.99762) (1.09565,0.99434) (1.09967,0.99041) (1.10368,0.9952) (1.10769,1.00166) (1.11171,1.00892) (1.11572,1.01699) (1.11973,1.02582) (1.12375,1.0354) (1.12776,1.04569) (1.13177,1.05665) (1.13579,1.06823) (1.1398,1.0804) (1.14381,1.09309) (1.14783,1.10627) (1.15184,1.11986)};
        \addplot[green!60, name path=lower, draw=none, forget plot] coordinates {(0.85084,0.73349) (0.85485,0.74894) (0.85886,0.76425) (0.86288,0.77936) (0.86689,0.79421) (0.8709,0.80875) (0.87492,0.82292) (0.87893,0.83666) (0.88294,0.84994) (0.88696,0.86269) (0.89097,0.87487) (0.89498,0.88644) (0.899,0.89736) (0.90301,0.89951) (0.90702,0.89961) (0.91104,0.9005) (0.91505,0.90217) (0.91906,0.90456) (0.92308,0.90763) (0.92709,0.9113) (0.9311,0.91551) (0.93512,0.9202) (0.93913,0.92528) (0.94314,0.93067) (0.94716,0.93629) (0.95117,0.94205) (0.95518,0.94786) (0.9592,0.95364) (0.96321,0.9593) (0.96722,0.96477) (0.97124,0.96995) (0.97525,0.97479) (0.97926,0.97919) (0.98328,0.98002) (0.98729,0.98009) (0.9913,0.98021) (0.99532,0.98045) (0.99933,0.98083) (1.00334,0.98138) (1.00736,0.98213) (1.01137,0.98309) (1.01538,0.98425) (1.0194,0.9856) (1.02341,0.98714) (1.02742,0.98883) (1.03144,0.98924) (1.03545,0.98709) (1.03946,0.98479) (1.04348,0.98245) (1.04749,0.98016) (1.05151,0.978) (1.05552,0.97605) (1.05953,0.97442) (1.06355,0.97316) (1.06756,0.97237) (1.07157,0.97211) (1.07559,0.97244) (1.0796,0.97342) (1.08361,0.97511) (1.08763,0.97754) (1.09164,0.98074) (1.09565,0.98474) (1.09967,0.98952) (1.10368,0.98569) (1.10769,0.98032) (1.11171,0.97425) (1.11572,0.96749) (1.11973,0.96004) (1.12375,0.95193) (1.12776,0.94318) (1.13177,0.93381) (1.13579,0.92386) (1.1398,0.91337) (1.14381,0.90237) (1.14783,0.8909) (1.15184,0.87901)};
        \addplot[green!60, opacity=1] fill between[of=upper and lower];
        \addlegendentry{95\% CI}  
        \end{axis}
    \end{tikzpicture}
    \caption{Improved GP approximation near the breakpoint}
\end{figure}

A way to circumvent the non-Gaussian behavior is to ensure that enough scenarios are simulated close to the breakpoint. This is unfortunately not guaranteed by the workflow \cite{Houdouin2024Certification}. If the first few simulated scenarios happen to be all on the same side of the breakpoint, a GP overconfidence problem arises and prevents any other simulations from being performed. If the first simulations do not reveal the non-linearity, the GP generalizes a fully linear behavior, cf. Fig. 5, for every possible scenario, and with enough confidence to replace the simulator. For all subsequent iterations, no other simulations are thus performed. For all scenarios $x \geq 1$, the GP confidently but inaccurately predicts congestion, ignoring the RES production limitation orders of the MPC. A wrong failure probability is thus estimated, leading to an erroneous certification. To avoid such undesirable results, the non-linearity induced by the MPC actions must be discovered.

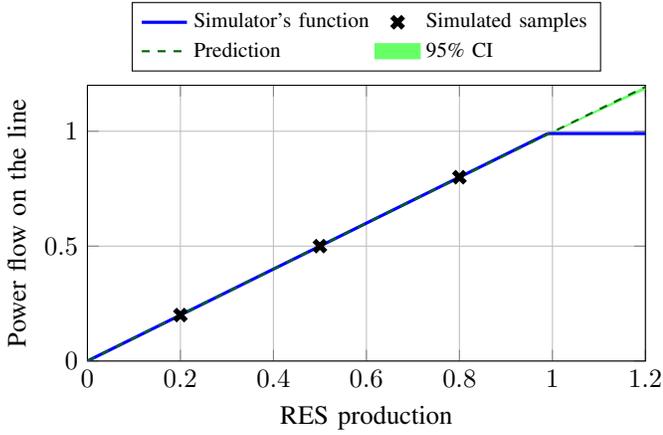
\begin{figure}[htbp]
    \centering
    \begin{tikzpicture}
        \begin{axis}[width=9cm, height=5.25cm, xmin=0, xmax=1.2, ymin=0, ymax=1.2, xlabel={RES production}, ylabel={Power flow on the line}, grid=major, legend style={at={(0.5,1.05)}, anchor=south, legend columns=2, font=\footnotesize}, legend cell align={left}]
        \addplot [blue, very thick, domain=0:0.99] {x};
        \addlegendentry{Simulator's function}
        \addplot [blue, very thick, domain=0.99:1.2, forget plot] {0.99};
        \addplot[only marks, mark=x, line width=2pt, mark size=3pt, black] coordinates {(0.2, 0.2) (0.5, 0.5) (0.8, 0.8)};
        \addlegendentry{Simulated samples}
        \addplot[green!40!black, dashed, thick] coordinates {(0.00000,0.00035) (0.01212,0.01242) (0.02424,0.02449) (0.03636,0.03657) (0.04848,0.04865) (0.06061,0.06073) (0.07273,0.07282) (0.08485,0.08491) (0.09697,0.09701) (0.10909,0.10911) (0.12121,0.12121) (0.13333,0.13331) (0.14545,0.14542) (0.15758,0.15753) (0.16970,0.16964) (0.18182,0.18176) (0.19394,0.19387) (0.20606,0.20599) (0.21818,0.21811) (0.23030,0.23023) (0.24242,0.24236) (0.25455,0.25448) (0.26667,0.26661) (0.27879,0.27874) (0.29091,0.29086) (0.30303,0.30299) (0.31515,0.31512) (0.32727,0.32725) (0.33939,0.33938) (0.35152,0.35151) (0.36364,0.36363) (0.37576,0.37576) (0.38788,0.38789) (0.40000,0.40002) (0.41212,0.41214) (0.42424,0.42427) (0.43636,0.43639) (0.44848,0.44851) (0.46061,0.46063) (0.47273,0.47275) (0.48485,0.48487) (0.49697,0.49698) (0.50909,0.50909) (0.52121,0.52120) (0.53333,0.53331) (0.54545,0.54542) (0.55758,0.55752) (0.56970,0.56962) (0.58182,0.58171) (0.59394,0.59380) (0.60606,0.60589) (0.61818,0.61798) (0.63030,0.63006) (0.64242,0.64213) (0.65455,0.65420) (0.66667,0.66627) (0.67879,0.67833) (0.69091,0.69039) (0.70303,0.70244) (0.71515,0.71449) (0.72727,0.72654) (0.73939,0.73858) (0.75152,0.75062) (0.76364,0.76266) (0.77576,0.77469) (0.78788,0.78671) (0.80000,0.79873) (0.81212,0.81075) (0.82424,0.82276) (0.83636,0.83477) (0.84848,0.84677) (0.86061,0.85877) (0.87273,0.87077) (0.88485,0.88276) (0.89697,0.89475) (0.90909,0.90673) (0.92121,0.91871) (0.93333,0.93068) (0.94545,0.94265) (0.95758,0.95461) (0.96970,0.96657) (0.98182,0.97852) (0.99394,0.99047) (1.00606,1.00241) (1.01818,1.01435) (1.03030,1.02629) (1.04242,1.03822) (1.05455,1.05015) (1.06667,1.06207) (1.07879,1.07399) (1.09091,1.08591) (1.10303,1.09782) (1.11515,1.10973) (1.12727,1.12163) (1.13939,1.13353) (1.15152,1.14543) (1.16364,1.15732) (1.17576,1.16921) (1.18788,1.18109) (1.20000,1.19297)};
        \addlegendentry{Prediction}
        \addplot[green!60, name path=upper, draw=none, forget plot] coordinates {(0.00000,0.00085) (0.01212,0.01287) (0.02424,0.02490) (0.03636,0.03694) (0.04848,0.04899) (0.06061,0.06104) (0.07273,0.07310) (0.08485,0.08517) (0.09697,0.09724) (0.10909,0.10932) (0.12121,0.12141) (0.13333,0.13350) (0.14545,0.14561) (0.15758,0.15771) (0.16970,0.16982) (0.18182,0.18194) (0.19394,0.19406) (0.20606,0.20619) (0.21818,0.21831) (0.23030,0.23044) (0.24242,0.24257) (0.25455,0.25469) (0.26667,0.26682) (0.27879,0.27895) (0.29091,0.29108) (0.30303,0.30321) (0.31515,0.31534) (0.32727,0.32746) (0.33939,0.33959) (0.35152,0.35172) (0.36364,0.36384) (0.37576,0.37596) (0.38788,0.38809) (0.40000,0.40021) (0.41212,0.41233) (0.42424,0.42445) (0.43636,0.43657) (0.44848,0.44870) (0.46061,0.46082) (0.47273,0.47295) (0.48485,0.48508) (0.49697,0.49721) (0.50909,0.50934) (0.52121,0.52147) (0.53333,0.53361) (0.54545,0.54574) (0.55758,0.55788) (0.56970,0.57001) (0.58182,0.58214) (0.59394,0.59428) (0.60606,0.60641) (0.61818,0.61854) (0.63030,0.63067) (0.64242,0.64280) (0.65455,0.65492) (0.66667,0.66704) (0.67879,0.67916) (0.69091,0.69128) (0.70303,0.70340) (0.71515,0.71551) (0.72727,0.72762) (0.73939,0.73972) (0.75152,0.75182) (0.76364,0.76392) (0.77576,0.77602) (0.78788,0.78811) (0.80000,0.80019) (0.81212,0.81227) (0.82424,0.82435) (0.83636,0.83642) (0.84848,0.84849) (0.86061,0.86055) (0.87273,0.87261) (0.88485,0.88466) (0.89697,0.89671) (0.90909,0.90875) (0.92121,0.92079) (0.93333,0.93282) (0.94545,0.94484) (0.95758,0.95686) (0.96970,0.96887) (0.98182,0.98087) (0.99394,0.99287) (1.00606,1.00486) (1.01818,1.01684) (1.03030,1.02882) (1.04242,1.04079) (1.05455,1.05275) (1.06667,1.06470) (1.07879,1.07665) (1.09091,1.08859) (1.10303,1.10052) (1.11515,1.11244) (1.12727,1.12435) (1.13939,1.13626) (1.15152,1.14815) (1.16364,1.16004) (1.17576,1.17192) (1.18788,1.18379) (1.20000,1.19565)};
        \addplot[green!60, name path=lower, draw=none, forget plot] coordinates {(0.00000,-0.00015) (0.01212,0.01196) (0.02424,0.02407) (0.03636,0.03619) (0.04848,0.04830) (0.06061,0.06042) (0.07273,0.07254) (0.08485,0.08466) (0.09697,0.09677) (0.10909,0.10889) (0.12121,0.12100) (0.13333,0.13312) (0.14545,0.14523) (0.15758,0.15735) (0.16970,0.16946) (0.18182,0.18157) (0.19394,0.19368) (0.20606,0.20580) (0.21818,0.21791) (0.23030,0.23003) (0.24242,0.24215) (0.25455,0.25427) (0.26667,0.26639) (0.27879,0.27852) (0.29091,0.29064) (0.30303,0.30277) (0.31515,0.31490) (0.32727,0.32703) (0.33939,0.33916) (0.35152,0.35130) (0.36364,0.36343) (0.37576,0.37556) (0.38788,0.38769) (0.40000,0.39983) (0.41212,0.41196) (0.42424,0.42408) (0.43636,0.43621) (0.44848,0.44835) (0.46061,0.46048) (0.47273,0.47260) (0.48485,0.48473) (0.49697,0.49676) (0.50909,0.50885) (0.52121,0.52094) (0.53333,0.53302) (0.54545,0.54509) (0.55758,0.55716) (0.56970,0.56922) (0.58182,0.58128) (0.59394,0.59333) (0.60606,0.60537) (0.61818,0.61741) (0.63030,0.62944) (0.64242,0.64146) (0.65455,0.65348) (0.66667,0.66548) (0.67879,0.67748) (0.69091,0.68947) (0.70303,0.70144) (0.71515,0.71341) (0.72727,0.72536) (0.73939,0.73731) (0.75152,0.74924) (0.76364,0.76116) (0.77576,0.77307) (0.78788,0.78496) (0.80000,0.79685) (0.81212,0.80872) (0.82424,0.82058) (0.83636,0.83242) (0.84848,0.84425) (0.86061,0.85607) (0.87273,0.86788) (0.88485,0.87967) (0.89697,0.89145) (0.90909,0.90321) (0.92121,0.91496) (0.93333,0.92669) (0.94545,0.93841) (0.95758,0.95011) (0.96970,0.96179) (0.98182,0.97346) (0.99394,0.98511) (1.00606,0.99675) (1.01818,1.00837) (1.03030,1.01997) (1.04242,1.03155) (1.05455,1.04312) (1.06667,1.05467) (1.07879,1.06620) (1.09091,1.07771) (1.10303,1.08921) (1.11515,1.10068) (1.12727,1.11214) (1.13939,1.12358) (1.15152,1.13501) (1.16364,1.14641) (1.17576,1.15780) (1.18788,1.16917) (1.20000,1.18052)};
        \addplot[green!60, opacity=1] fill between[of=upper and lower];
        \addlegendentry{95\% CI}  
        \end{axis}
    \end{tikzpicture}
    \caption{GP linear behavior generalization problem}
\end{figure}

\subsection{Adaptive residual uncertainty}

To ensure the non-linearity discovery, a first solution could be to randomly simulate scenarios, even if the GP is sufficiently confident about the outcome. At some point, scenarios near the breakpoint will be simulated, revealing the non-linearity. The only downside is that most randomly performed simulations will be useless. For most scenarios, indeed, the absence of congestion can be accurately predicted simply because there is not enough RES production to overload the lines, regardless of the non-linearity discovery. Enforcing the simulation of these harmless scenarios is a waste of time.

Another idea to enforce interesting simulations is to reduce the GP's confidence so that it replaces the simulator less often. This can be achieved by increasing the posterior uncertainty $\sigma_*$ by a residual uncertainty term $\sigma_{\text{ru}}$ and obtain the posterior distribution $\mathcal{N} \left( \mu_*, \sigma_* + \sigma_{\text{ru}} \right)$. For harmless scenarios far away from overloading the lines, the GP has a strong enough certainty about the outcome to remain sufficiently confident to replace the simulator despite the uncertainty increase. For scenarios closer to the breakpoint, the GP is no longer confident enough, cf. Fig.6, and a simulation is performed. Adding this residual uncertainty term thus enforces only useful simulations near the breakpoint, enabling the GP to remain reliable, like in Fig.4, despite the non-linearity.

\begin{figure}[htbp]
    \centering
    \begin{tikzpicture}
        \begin{axis}[width=9cm, height=5.25cm, xmin=0, xmax=1.2, ymin=0, ymax=1.2, xlabel={RES production}, ylabel={Power flow on the line}, grid=major, title style={at={(0.5, -0.15)}, anchor=north}, legend style={at={(0.5,1.05)}, anchor=south, legend columns=2, font=\footnotesize}, legend cell align={left}]
        \addplot [blue, very thick, domain=0:0.99] {x};
        \addlegendentry{Simulator's function}
        \addplot [blue, very thick, domain=0.99:1.2, forget plot] {0.99};
        \addplot[only marks, mark=x, line width=2pt, mark size=3pt, black] coordinates {(0.2, 0.2) (0.5, 0.5) (0.8, 0.8)};
        \addlegendentry{Simulated samples}
        \addplot[green!40!black, dashed, thick] coordinates {(0.00000,0.00035) (0.01212,0.01242) (0.02424,0.02449) (0.03636,0.03657) (0.04848,0.04865) (0.06061,0.06073) (0.07273,0.07282) (0.08485,0.08491) (0.09697,0.09701) (0.10909,0.10911) (0.12121,0.12121) (0.13333,0.13331) (0.14545,0.14542) (0.15758,0.15753) (0.16970,0.16964) (0.18182,0.18176) (0.19394,0.19387) (0.20606,0.20599) (0.21818,0.21811) (0.23030,0.23023) (0.24242,0.24236) (0.25455,0.25448) (0.26667,0.26661) (0.27879,0.27874) (0.29091,0.29086) (0.30303,0.30299) (0.31515,0.31512) (0.32727,0.32725) (0.33939,0.33938) (0.35152,0.35151) (0.36364,0.36363) (0.37576,0.37576) (0.38788,0.38789) (0.40000,0.40002) (0.41212,0.41214) (0.42424,0.42427) (0.43636,0.43639) (0.44848,0.44851) (0.46061,0.46063) (0.47273,0.47275) (0.48485,0.48487) (0.49697,0.49698) (0.50909,0.50909) (0.52121,0.52120) (0.53333,0.53331) (0.54545,0.54542) (0.55758,0.55752) (0.56970,0.56962) (0.58182,0.58171) (0.59394,0.59380) (0.60606,0.60589) (0.61818,0.61798) (0.63030,0.63006) (0.64242,0.64213) (0.65455,0.65420) (0.66667,0.66627) (0.67879,0.67833) (0.69091,0.69039) (0.70303,0.70244) (0.71515,0.71449) (0.72727,0.72654) (0.73939,0.73858) (0.75152,0.75062) (0.76364,0.76266) (0.77576,0.77469) (0.78788,0.78671) (0.80000,0.79873) (0.81212,0.81075) (0.82424,0.82276) (0.83636,0.83477) (0.84848,0.84677) (0.86061,0.85877) (0.87273,0.87077) (0.88485,0.88276) (0.89697,0.89475) (0.90909,0.90673) (0.92121,0.91871) (0.93333,0.93068) (0.94545,0.94265) (0.95758,0.95461) (0.96970,0.96657) (0.98182,0.97852) (0.99394,0.99047) (1.00606,1.00241) (1.01818,1.01435) (1.03030,1.02629) (1.04242,1.03822) (1.05455,1.05015) (1.06667,1.06207) (1.07879,1.07399) (1.09091,1.08591) (1.10303,1.09782) (1.11515,1.10973) (1.12727,1.12163) (1.13939,1.13353) (1.15152,1.14543) (1.16364,1.15732) (1.17576,1.16921) (1.18788,1.18109) (1.20000,1.19297)};
        \addlegendentry{Prediction}
        \addplot[green!60, name path=upper, draw=none, forget plot] coordinates {(0.00000,0.04935) (0.01212,0.06142) (0.02424,0.07349) (0.03636,0.08557) (0.04848,0.09765) (0.06061,0.10973) (0.07273,0.12182) (0.08485,0.13391) (0.09697,0.14601) (0.10909,0.15811) (0.12121,0.17021) (0.13333,0.18231) (0.14545,0.19442) (0.15758,0.20653) (0.16970,0.21864) (0.18182,0.23076) (0.19394,0.24287) (0.20606,0.25499) (0.21818,0.26711) (0.23030,0.27923) (0.24242,0.29136) (0.25455,0.30348) (0.26667,0.31561) (0.27879,0.32774) (0.29091,0.33986) (0.30303,0.35199) (0.31515,0.36412) (0.32727,0.37625) (0.33939,0.38838) (0.35152,0.40051) (0.36364,0.41263) (0.37576,0.42476) (0.38788,0.43689) (0.40000,0.44902) (0.41212,0.46114) (0.42424,0.47327) (0.43636,0.48539) (0.44848,0.49751) (0.46061,0.50963) (0.47273,0.52175) (0.48485,0.53387) (0.49697,0.54598) (0.50909,0.55809) (0.52121,0.57020) (0.53333,0.58231) (0.54545,0.59442) (0.55758,0.60652) (0.56970,0.61862) (0.58182,0.63071) (0.59394,0.64281) (0.60606,0.65489) (0.61818,0.66698) (0.63030,0.67906) (0.64242,0.69114) (0.65455,0.70321) (0.66667,0.71528) (0.67879,0.72734) (0.69091,0.73940) (0.70303,0.75146) (0.71515,0.76351) (0.72727,0.77555) (0.73939,0.78759) (0.75152,0.79962) (0.76364,0.81165) (0.77576,0.82367) (0.78788,0.83569) (0.80000,0.84770) (0.81212,0.85971) (0.82424,0.87170) (0.83636,0.88370) (0.84848,0.89568) (0.86061,0.90766) (0.87273,0.91963) (0.88485,0.93160) (0.89697,0.94355) (0.90909,0.95550)};
        \addplot[green!60, name path=lower, draw=none, forget plot] coordinates {(0.00000,-0.04865) (0.01212,-0.03659) (0.02424,-0.02451) (0.03636,-0.01244) (0.04848,-0.00036) (0.06061,0.01173) (0.07273,0.02382) (0.08485,0.03591) (0.09697,0.04801) (0.10909,0.06010) (0.12121,0.07221) (0.13333,0.08431) (0.14545,0.09642) (0.15758,0.10853) (0.16970,0.12064) (0.18182,0.13276) (0.19394,0.14487) (0.20606,0.15699) (0.21818,0.16911) (0.23030,0.18123) (0.24242,0.19336) (0.25455,0.20548) (0.26667,0.21761) (0.27879,0.22973) (0.29091,0.24186) (0.30303,0.25399) (0.31515,0.26612) (0.32727,0.27825) (0.33939,0.29038) (0.35152,0.30251) (0.36364,0.31464) (0.37576,0.32677) (0.38788,0.33890) (0.40000,0.35102) (0.41212,0.36315) (0.42424,0.37527) (0.43636,0.38740) (0.44848,0.39952) (0.46061,0.41164) (0.47273,0.42376) (0.48485,0.43587) (0.49697,0.44799) (0.50909,0.46010) (0.52121,0.47222) (0.53333,0.48433) (0.54545,0.49643) (0.55758,0.50854) (0.56970,0.52064) (0.58182,0.53274) (0.59394,0.54483) (0.60606,0.55692) (0.61818,0.56900) (0.63030,0.58107) (0.64242,0.59314) (0.65455,0.60519) (0.66667,0.61724) (0.67879,0.62928) (0.69091,0.64131) (0.70303,0.65333) (0.71515,0.66534) (0.72727,0.67734) (0.73939,0.68932) (0.75152,0.70130) (0.76364,0.71326) (0.77576,0.72521) (0.78788,0.73714) (0.80000,0.74906) (0.81212,0.76097) (0.82424,0.77286) (0.83636,0.78474) (0.84848,0.79660) (0.86061,0.80845) (0.87273,0.82028) (0.88485,0.83210) (0.89697,0.84390) (0.90909,0.85568)};
        \addplot[green!60, opacity=1] fill between[of=upper and lower];
        \addlegendentry{95\% CI}   
        \addplot[green!60, name path=upper, draw=none, forget plot] coordinates {(1.10303,1.14557) (1.11515,1.15738) (1.12727,1.16917) (1.13939,1.18095) (1.15152,1.19272) (1.16364,1.20448) (1.17576,1.21623) (1.18788,1.22798) (1.20000,1.23970)};
        \addplot[green!60, name path=lower, draw=none, forget plot] coordinates {(1.10303,1.04193) (1.11515,1.05342) (1.12727,1.06489) (1.13939,1.07634) (1.15152,1.08777) (1.16364,1.09918) (1.17576,1.11057) (1.18788,1.12194) (1.20000,1.13329)};
        \addplot[green!60, opacity=1, forget plot] fill between[of=upper and lower];
        \addplot[orange!60, name path=upper, draw=none, forget plot] coordinates {(0.90909,0.95550) (0.92121,0.96744) (0.93333,0.97938) (0.94545,0.99131) (0.95758,1.00322) (0.96970,1.01513) (0.98182,1.02704) (0.99394,1.03893) (1.00606,1.05082) (1.01818,1.06269) (1.03030,1.07456) (1.04242,1.08642) (1.05455,1.09827) (1.06667,1.11011) (1.07879,1.12194) (1.09091,1.13376) (1.10303,1.14557)};
        \addplot[orange!60, name path=lower, draw=none, forget plot] coordinates {(0.90909,0.85568) (0.92121,0.86745) (0.93333,0.87920) (0.94545,0.89094) (0.95758,0.90266) (0.96970,0.91436) (0.98182,0.92605) (0.99394,0.93772) (1.00606,0.94937) (1.01818,0.96100) (1.03030,0.97262) (1.04242,0.98421) (1.05455,0.99579) (1.06667,1.00735) (1.07879,1.01890) (1.09091,1.03042) (1.10303,1.04193)};
        \addplot[orange!60, opacity=1] fill between[of=upper and lower];
        \addlegendentry{GP no longer confident enough} 
        \end{axis}
    \end{tikzpicture}
    \caption{Residual uncertainty to enforce informative simulations}
\end{figure}
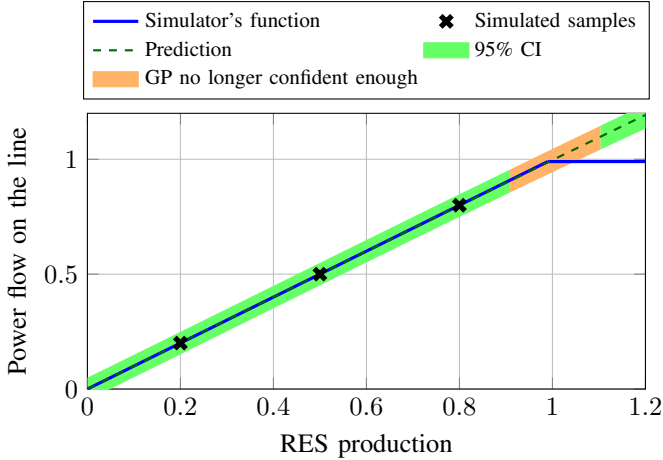

However, this early advantage for the GP is also an asymptotic drawback for the workflow. The residual uncertainty creates a region close to the breakpoint where the GP cannot be confident enough to replace the simulator. It corresponds to all predictions where $|1 - \mu_*| < \sqrt{2} \sigma_{\text{ru}} erf^{-1}(1-2\beta)$. A possibility to preserve the workflow's asymptotic performances is to adapt the residual uncertainty with the workflow's progress. The idea is to add a large residual uncertainty at the beginning of the workflow to discover the non-Gaussian behaviors and then gradually decrease it when they have been properly characterized. The goal is to end up with no residual uncertainty so that the GP can asymptotically avoid all simulations. In this paper, a simple exponential decrease strategy is used: $\sigma_{\text{ru}}(n) = \frac{\sigma_{\text{ru}}^0}{\alpha^n}$. Hyperparameters $\sigma_{\text{ru}}^0$ and $\alpha$ are crucial as they directly control the trade-off between GP discovery and workflow asymptotic efficiency.

\section{Computational results}

The proposed workflow enhancement is applied to a real zone of the French transmission grid called Jalancourt (cf. Fig.7), located in Burgundy. This zone includes 15 lines and 13 RES production units, which is small compared to others that contain between tens and a hundred lines and nodes. A scenario is thus represented by a vector $\boldsymbol{x} \in \mathbb{R}^{13}$. Power flows in the lines are computed using the load flow equations $y = M (\boldsymbol{x} - \boldsymbol{\Delta x})$, where $\boldsymbol{\Delta x}$ denotes the amount of production curtailed by the MPC. The MPC limitation vector $\boldsymbol{\Delta x}$ is determined by solving an optimization problem that minimizes the amount of RES production curtailed while ensuring that line flows remain below the threshold $F_{\text{max}}$. Since $F_{\text{max}}$ is set to 0.99 for all scenarios, the MPC always successfully avoids congestion in all cases, and $p_{\text{failure}}=0$. 

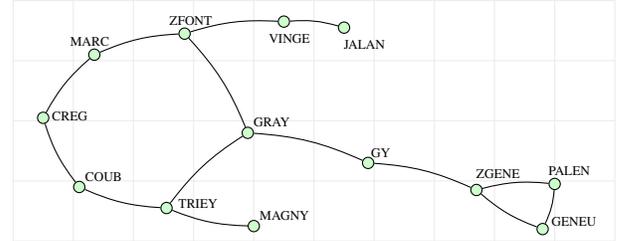
\begin{figure}[htbp]
    \centering
\begin{tikzpicture}[scale=4, font=\tiny, background grid/.style={step=0.2cm, gray!15}, node circle/.style={circle, draw, minimum size=0.15cm, inner sep=0pt, fill=green!20}, node label/.style={inner sep=0pt, outer sep=0pt, draw=none}, every path/.style={draw, rounded corners}]

\draw[background grid] (-0.8,-0.4) grid (1.2,0.4);

\node[node circle] (MARC) at (-0.53,0.22) {};
\node[node label, anchor=south west] at (-0.61,0.25) {MARC};

\node[node circle] (GRAY) at (-0.02,-0.04) {};
\node[node label, anchor=south west] at (0,-0.02) {GRAY};

\node[node circle] (GENEU) at (0.96,-0.36) {};
\node[node label, anchor=south west] at (0.99,-0.35) {GENEU};

\node[node circle] (GY) at (0.38,-0.14) {};
\node[node label, anchor=south west] at (0.39,-0.12) {GY};

\node[node circle] (ZFONT) at (-0.23,0.29) {};
\node[node label, anchor=south west] at (-0.28,0.32) {ZFONT};

\node[node circle] (CREG) at (-0.70,0.01) {};
\node[node label, anchor=south west] at (-0.67,0.0) {CREG};

\node[node circle] (VINGE) at (0.1,0.33) {};
\node[node label, anchor=south west] at (0.05,0.26) {VINGE};

\node[node circle] (ZGENE) at (0.74,-0.23) {};
\node[node label, anchor=south west] at (0.74,-0.19) {ZGENE};

\node[node circle] (MAGNY) at (0,-0.35) {};
\node[node label, anchor=south west] at (0.02,-0.33) {MAGNY};

\node[node circle] (TRIEY) at (-0.29,-0.29) {};
\node[node label, anchor=south west] at (-0.25,-0.29) {TRIEY};

\node[node circle] (PALEN) at (1.00,-0.21) {};
\node[node label, anchor=south west] at (0.98,-0.18) {PALEN};

\node[node circle] (JALAN) at (0.3,0.31) {};
\node[node label, anchor=south west] at (0.30,0.24) {JALAN};

\node[node circle] (COUB) at (-0.58,-0.22) {};
\node[node label, anchor=south west] at (-0.56,-0.2) {COUB};

\draw[bend right=10] (MARC) to (CREG);
\draw[bend left=10] (MARC) to (ZFONT);
\draw[bend left=10] (GRAY) to (GY);
\draw[bend right=10] (GRAY) to (TRIEY);
\draw[bend right=10] (GRAY) to (ZFONT);
\draw[bend right=10] (GENEU) to (PALEN);
\draw[bend left=10] (GENEU) to (ZGENE);
\draw[bend left=10] (GY) to (ZGENE);
\draw[bend left=10] (ZFONT) to (VINGE);
\draw[bend right=10] (CREG) to (COUB);
\draw[bend left=10] (VINGE) to (JALAN);
\draw[bend left=10] (ZGENE) to (PALEN);
\draw[bend left=10] (MAGNY) to (TRIEY);
\draw[bend left=10] (TRIEY) to (COUB);

\end{tikzpicture}
    \caption{Jalancourt Zone}
\end{figure}

In these experiments, the number of scenarios used to estimate the failure probability is set to $N_{\text{scenario}}=2000$. An effective solution should provide both an accurate estimation of the failure probability and perform as few simulations as possible. The performance of the proposed method with adaptive residual uncertainty (ARU) is compared to two other state-of-the-art surrogate-model-based approaches: Latin Hypercube Sampling (LHS) \cite{Mckay1979Comparison} and a Bayesian strategy \cite{Moss2024Bayesian}. Unlike the proposed workflow \cite{Houdouin2024Certification}, these two methods first train a GP as efficiently as possible, then use the obtained model to estimate the failure probability without performing additional simulations. Following recommendations from the literature, the prior number of simulations for LHS and Bayesian is set to $N_{\text{prior simulation}} = 10 \times d = 130$. Hyperparameters used for the adaptive residual uncertainty strategy are $\sigma_{\text{ru}}^0=0.1$ and $\alpha=1.2$, and the hyperparameter for the proxy use decision is set to $\beta = 0.01$.

\begin{table}[htbp]
\centering
\renewcommand{\arraystretch}{1.4}
\begin{tabular}{|p{1.1cm}|p{1.7cm}|p{2cm}|p{2cm}|}
\hline
\textbf{Method} & \textbf{Estimation of $p_{\text{failure}}$} & \textbf{\% of performed simulations} & \textbf{\% of misclassified scenarios} \\
\hline
\textbf{LHS} & $0.081 \pm 0.012$ & 6.5\% & 12.80\% \\
\hline
\textbf{Bayesian} & $0.009 \pm 0.004$ & 6.5\% & 4.65\% \\
\hline
\textbf{ARU} & $0.003 \pm 0.001$ & 10.3\% & 0.34\% \\
\hline
\end{tabular}
\caption{Experimental results on Jalancourt zone}
\end{table}

These results are provided for illustrative purposes, as all methods can achieve accurate estimations given a sufficient number of simulated scenarios. The key distinction of the proposed method compared to state-of-the-art approaches lies in its simplicity: it does not require assessing whether the GP has enough training samples to correctly and reliably approximate the simulator, and provide an accurate failure probability estimation. On the other hand, unlike with the Bayesian approach, the simulated scenarios may not be optimally located in order to minimize the total number of simulations to perform. There is thus a trade-off between sample efficiency and the uncertainty regarding the number of simulations to perform.

\section{Conclusions and perspectives}

This work aims to certify the proper functioning of an MPC algorithm responsible for managing power grid congestion. The certification is achieved by testing the ability of the MPC to properly manage congestion in various scenarios. If all scenarios are simulated with a realistic, and thus computationally intensive, simulator, the workflow cannot terminate in a reasonable amount of time. A surrogate model of the simulator is thus learned in parallel with the certification to avoid some simulations and speed up the workflow.

Among the various ML models commonly used to build the surrogate model, GPs are a very popular choice. GPs are flexible, data-efficient, and interpretable ML models that jointly provide prediction and uncertainty quantification. Uncertainty quantification enables using the surrogate model only when it is confident enough about the outcome, and limits the chance of trusting a flawed model. GPs, however, fail to learn non-smooth behaviors, such as those exhibited by the network simulator. To preserve the reliability of the surrogate model, a new approach is proposed in which a residual uncertainty is added to the posterior distribution to account for potential non-Gaussian behaviors. The residual uncertainty enforces more simulations in the early stage of the workflow, which provides enough data to the GP to accurately and reliably characterize the non-linear behavior of the simulator. The residual uncertainty is then gradually decreased to allow the GP to reach sufficient confidence to replace the simulator for every possible scenario. The current strategy is an exponential decay of the residual uncertainty, but more complex adaptive strategies could be considered. The proposed approach is tested on a real zone of the French transmission grid with a simple network simulator. It accurately and reliably approximates the simulator after a few tens of iterations and enables the avoidance of all the simulations asymptotically, achieving the goal of considerably accelerating the certification workflow. For now, only experiments involving simple and thus fast-to-evaluate simulators were successfully conducted. This allows future work to focus on more complex experiments with realistic but time-consuming simulators, making the benefit of GP concrete.

Ensuring a reliable uncertainty quantification despite non-Gaussian behaviors is a challenge in the GP literature. The chosen approach aims to add a constantly decreasing variance term to enlarge the confidence intervals and afford some irregularities in the function's behavior. More complex works look to overcome this limitation by mixing GPs with model-free UQ techniques such as conformal predictions \cite{Jaber2024Conformal}. It is another possibility to explore to enable more reliable confidence intervals.

\end{document}